\documentclass[journal]{IEEEtran}

\usepackage{times,amsmath,amssymb}
\usepackage{graphicx}
\usepackage{xcolor}
\usepackage{setspace}

\usepackage{multirow}
\usepackage[noend]{algpseudocode}

\usepackage{caption}
{}
\usepackage[ruled,vlined]{algorithm2e}{}
{}

\newenvironment{pseudocode}[1][htb]
  {% Update algorithm name
   \begin{algorithm}[#1]%
  }{\end{algorithm}}

\begin{document}

%\supertitle{Neural Class-Specific Regression for face verification}
\title{Neural Class-Specific Regression for face verification}
\author{{Guanqun Cao}, {Alexandros Iosifidis}, {Moncef Gabbouj}}
%\address{$^{\dag}${Laboratory of Signal Processing, Tampere University of Technology, Tampere, Finland}\\
%$^{\ddag}${Department of Engineering, Electrical and Computer Engineering, Aarhus University, Aarhus, Denmark}\\
%\email{\{guanqun.cao,moncef.gabbouj\}@tut.fi}, alexandros.iosifidis@eng.au.dk}

\maketitle
\begin{abstract}
Face verification is a problem approached in the literature mainly using nonlinear class-specific subspace learning techniques. While it has been shown that kernel-based Class-Specific Discriminant Analysis is able to provide excellent performance in small- and medium-scale face verification problems, its application in today's large-scale problems is difficult due to its training space and computational requirements. In this paper, generalizing our previous work on kernel-based class-specific discriminant analysis, we show that class-specific subspace learning can be cast as a regression problem. This allows us to derive linear, (reduced) kernel and neural network-based class-specific discriminant analysis methods using efficient batch and/or iterative training schemes, suited for large-scale learning problems. We test the performance of these methods in two datasets describing medium- and large-scale face verification problems.
\end{abstract}

\section{Introduction}\label{S:Introduction}
\begin{figure*}[h!]
    \centering{\includegraphics[height=2.6in]{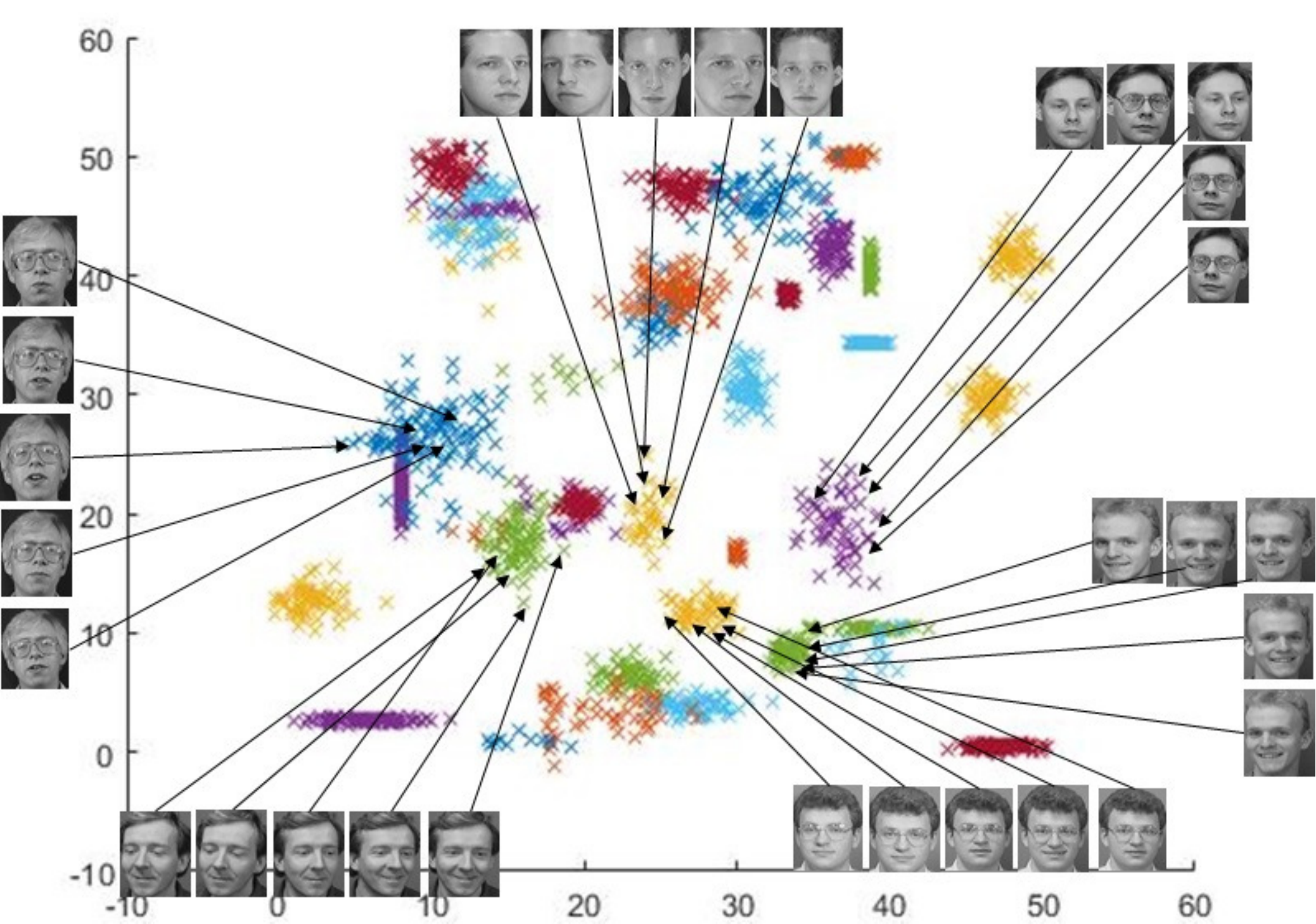}\includegraphics[height=2.6in]{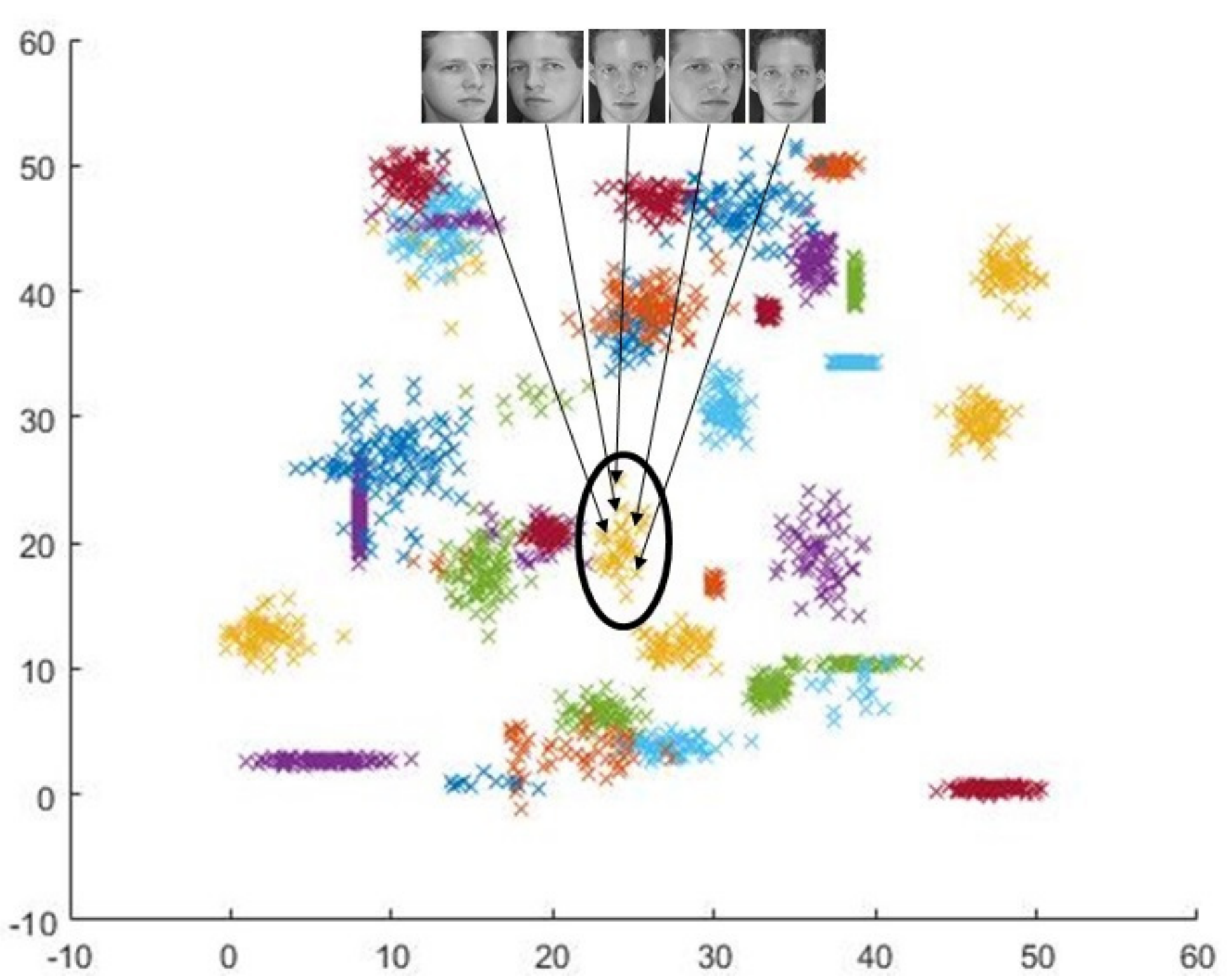}}                                                                                                                                  \caption{\it Illustration of face recognition and face verification problems. (Left) Face recognition is a multi-class problem, where given a new facial image the answer is the ID of the depicted person. (Right) Face verification is a binary problem, where given a new facial image the answer is whether the image depicts the ID of interest or not. Here we show the 2-D representations of the facial vectors in ORL dataset \cite{samaria1994parameterisation} obtained by applying PCA.}\label{fig:FaceRecognitionVerification}
\end{figure*}

Facial image analysis received intensive research attention during the last two decades, due to its importance in a wide variety of applications, ranging from surveillance, affective computing, entertainment and assisted living \cite{Barr2011video,Li2011discriminative,Iosifidis2012activity}. Depending on the application scenario, different facial image analysis problems are considered, the most widely used ones being those of face recognition and face verification. On the one hand, face recognition is a multi-class problem, where the objective is to categorize a new (unknown) facial image in one of the classes defined by all person IDs included in a facial image database. On the other hand, face verification is a binary problem, where the objective is to distinguish one class (usually called positive class) defined by the ID of the person of interest from the rest of the world (usually called negative class formed by the IDs of all other persons, who might not even be included in the facial image database). An illustration of the face recognition and face verification problems is shown in Figure \ref{fig:FaceRecognitionVerification}.

One line of work in face verification exploits the power of Subspace Learning techniques. While it has been shown that unsupervised subspace learning techniques, like Principal Component Analysis \cite{Duda2000}, Locally Linear Embedding \cite{Saul2003Think} and Locality Preserving Projections \cite{He2003Locality}, can successfully capture facial image manifolds, their unsupervised nature usually leads to lower performance compared to supervised subspace learning approaches. Perhaps the most well-known and commonly applied supervised subspace learning technique is Linear Discriminant Analysis (LDA) and its variants \cite{Duda2000,Ye2007least,Iosifidis2013LDA}. LDA (under the Gaussian class assumption) defines the optimal linear projection from the input space to the discriminant (sub)space where the within-class scatter is minimized, while the between-class scatter is maximized. Extensions of LDA exploiting kernels, like Kernel Discriminant Analysis (KDA) \cite{Baudat2000Generalized} and Kernel Reference Discriminant Analysis (KRDA) \cite{Iosifidis2014KRDA} can also define non-linear discriminant spaces, greatly enhancing performance in non-linear problems.

While LDA and its variants have shown to achieve very good performance in multi-class problems, like face recognition, their performance in face verification problems (which are usually defined as binary problems) is limited by the fact that the maximal dimensionality of the derived discriminant (sub-)space is restricted by the number of classes. This is a result of the within-class and between-class scatters definition, making the maximal rank of the corresponding matrices for a $P$-class problem equal to $P-1$. That is, for verification problems, the maximal discriminant (sub-)space dimensionality derived by LDA is equal to one. Class-Specific Discriminant Analysis (CSDA) techniques have been proposed to overcome this restriction by exploiting intra-class and out-of-class scatter definitions leading to matrices of higher ranks \cite{Kittler2000Face,Goudelis2007,Arashloo2014class,Iosifidis2015CSRDA}. As a consequence, class-specific techniques have been shown to outperform their multi-class counterparts in verification problems, exploiting data representations in discriminant (sub-)spaces of higher dimensionality.

Another issue that should be appropriately addressed, both for multi-class and class-specific approaches, is related to the space and computational costs of their non-linear versions based on kernels. For a training set formed by $N$ samples, standard kernel-based solutions require $O(N^2)$ storage size and $O(N^3)$ computations, rendering their application in today's large-scale problems difficult. Solutions based on low-rank approximations \cite{williams2001using,drineas2005nystrom} and reduced kernels \cite{Lee2007reduced,Iosifidis2016scalling} have been proposed in order to highly reduce both costs, while achieving satisfactory performance. In our previous work, we have shown that the non-linear version of CSDA based on kernels is equivalent to a kernel-regression problem and, thus, its computational cost can be reduced by exploiting efficient linear system solutions \cite{Iosifidis2017CSKDAr}. In addition, we have shown that eigenanalysis of the graph Laplacians defined in Class-Specific kernel Spectral Regression can be efficiently computed using a matrix factorization process taking into account the class labels of the training samples, leading to an efficient approximate CS-KDA solution \cite{Iosifidis2016scalling,Iosifidis2016IPTA}.

In this paper, we build on top of our previous work \cite{Iosifidis2017CSKDAr,Iosifidis2016scalling} and show that the linear and reduced kernel versions of CSDA are equivalent to a linear and a reduced kernel regression problem, respectively. Casting the linear CSDA criterion as a linear regression problem allows us to view CSDA as a processing block that can be used for iterative optimization on top of a (possibly deep) neural network topology. Based on that, we propose a non-linear CSDA solution based on neural networks. While neural network-based solutions for multi-class discriminant analysis have been recently proposed \cite{Wong2011deep,Stuhlsatz2012feature,Cao2016generalized}, this is the first time that neural networks are used for optimizing class-specific projections. We apply all three (linear, approximate kernel and neural) CSDA variants on two publicly available datasets describing medium- and large-scale face verification problems and compare their performance with related methods.

The paper is structured as follows. In Section \ref{S:ProblemStatement}, we provide an overview of the face verification problem. Linear and kernel-based CSDA techniques are briefly described in Section \ref{S:OriginalMethods}. We provide our analysis in Section \ref{S:Proposed}. We first show that the linear (subsection \ref{SS:ProposedLinear}) and reduced kernel (subsection \ref{SS:ProposedKernelCSLDA}) versions of CSDA are equivalent to regression problems using class-specific target vectors. Subsequently, we describe the proposed neural network-based CSDA in subsection \ref{SS:ProposedNetworkCSLDA}. Experiments on medium- and large-scale face verification problems are provided in Section \ref{S:Experiments} and conclusions are drawn in Section \ref{S:Conclusions}.

\section{Problem Statement}\label{S:ProblemStatement}
Let us assume that a facial image database is formed by $N$ images, each depicting a person belonging to an ID set $\mathcal{P} = \{1,\dots,p,\dots,P\}$. Let us also assume that these images have been pre-processed in order to produce the so-called facial image vectors $\mathbf{x}_i \in \mathbb{R}^D, \:i=1,\dots,N$. Vector $\mathbf{x}_i$ represents the $i$-th facial image in the database and is followed by an ID label $l_i \in \mathcal{P}$.

Given the above, we would like to determine a class-specific model discriminating person $p$ from all other persons. We will define this class-specific model by learning a (non-)linear mapping from the input space $\mathbb{R}^D$ to a low-dimensional (discriminant) space $\mathbb{R}^{d_p}, \:d_p \le D$, in which class $p$ is represented by the corresponding mean vector:
\begin{equation}
\bar{\mathbf{z}}_p = \frac{1}{N_p} \sum_{i, l_i = p} \mathbf{z}_i,
\end{equation}
where $N_p$ is the cardinality of class $p$ in the facial image database. $N_n = N - N_p$ denotes the cardinality of the negative class (formed by the facial images not belonging to class $p$). $\mathbf{z}_i = f(\mathbf{x}_i, \mathcal{W}_p)$ is the image of $\mathbf{x}_i$ in $\mathbb{R}^{d_p}$ obtained by optimizing the parameters $\mathcal{W}_p$ of function $f(\cdot)$ for achieving the maximal class-specific discrimination. After determining the (non-)linear mapping parameters $\mathcal{W}_p$ and the class mean vector $\bar{\mathbf{z}}_p$, a new facial image vector $\mathbf{z} \in \mathbb{R}^d$ calculated by $\mathbf{z} = f(\mathbf{x}, \mathcal{W}_p)$ should be close to $\bar{\mathbf{z}}_p$, if it depicts person $p$, or far from it, if it depicts another person.
\\ \\
\textbf{Notations:} We define by $\mathbf{e} \in \mathbb{R}^N$ a vector of ones, $\mathbf{e}_I \in \mathbb{R}^N$ a binary vector having elements $[\mathbf{e}_I]_i = 1$ if $l_i = p$ and $[\mathbf{e}_I]_i = 0$ if $l_i \neq p$ and $\mathbf{e}_O = \mathbf{e} - \mathbf{e}_I$. We also define the matrices $\mathbf{E}_I = \mathbf{e}_I \mathbf{e}_I^T$ and $\mathbf{E}_O = \mathbf{e}_O \mathbf{e}_O^T$. $\mathbf{X} \in \mathbb{R}^{D \times N}$ is a matrix formed by the facial vectors $\mathbf{x}_i$ as columns. %We also define by $E_I \in \mathbb{R}^{N \times N_p}$ and $E_O \in \mathbb{R}^{N \times N-N_p}$ two matrices having ones on their diagonal.

\section{Standard Class-Specific Discriminant Analysis}\label{S:OriginalMethods}
\begin{figure}[h!]
\centering \centerline{
\includegraphics[width=0.98\linewidth]{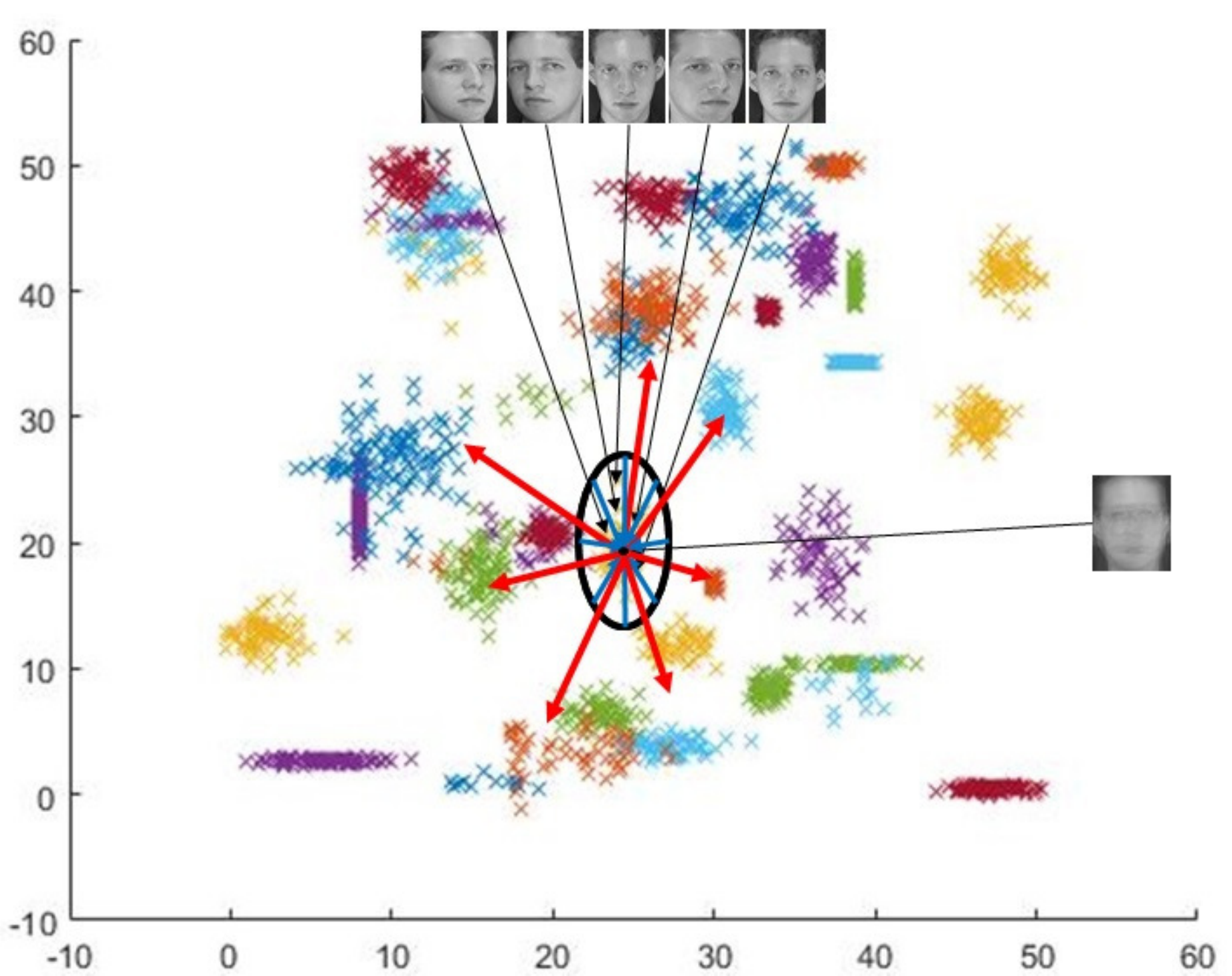} }
\caption{\it Face verification criterion. Facial images depicting the person of interest are forced to be as close as possible to the mean of the positive class (in this case the mean facial image depicted on the right), while facial images forming the negative class are forced to be as far as possible for it.}\label{fig:FaceVerificationCriterion}
\end{figure}

Let us denote by $D_I$ and $D_O$ the intra-class and out-of-class distances defined as:
\begin{equation}
D_I = \sum_{i, l_i = p} \| \mathbf{z}_i - \bar{\mathbf{z}}_p \|_2^2 = \sum_{i, l_i = p} \| f(\mathbf{x}_i, \mathcal{W}_p) - \bar{\mathbf{z}}_p \|_2^2
\end{equation}
and
\begin{equation}
D_O = \sum_{i, l_i \neq p} \| \mathbf{z}_i - \bar{\mathbf{z}}_p \|_2^2 = \sum_{i, l_i \neq p} \| f(\mathbf{x}_i, \mathcal{W}_p) - \bar{\mathbf{z}}_p \|_2^2.
\end{equation}
The parameters of the class-specific model $\mathcal{W}_p$ are optimized so that the intra-class distance is minimized and the out-of-class distance is maximized, as illustrated in Figure \ref{fig:FaceVerificationCriterion}. This can be expressed as maximizing the criterion:
\begin{equation}
\mathcal{J}(\mathcal{W}_p) = \frac{D_O(\mathcal{W}_p)}{D_I(\mathcal{W}_p)}.
\end{equation}

\subsection{Linear case}\label{SS:OriginalLinearCSLDA}
In the case where $f(\cdot)$ corresponds to a linear mapping \cite{Kittler2000Face}, $D_I$ and $D_O$ are given by:
\begin{equation}
D_I = \sum_{i, l_i = p} \| \mathbf{W}^T \mathbf{x}_i - \mathbf{W}^T \bar{\mathbf{x}}_p \|_2^2 = tr( \mathbf{W}^T \mathbf{S}_I \mathbf{W} )
\end{equation}
and
\begin{equation}
D_O = \sum_{i, l_i \neq p} \| \mathbf{W}^T \mathbf{x}_i - \mathbf{W}^T \bar{\mathbf{x}}_p \|_2^2 = tr( \mathbf{W}^T \mathbf{S}_O \mathbf{W} ),
\end{equation}
where $\mathbf{W} \in \mathbb{R}^{D \times d_p}$ is the projection matrix, linearly mapping the input space $\mathbb{R}^D$ to the discriminant subspace $\mathbb{R}^{d_p}$ and $\bar{\mathbf{x}}_p = \frac{1}{N_p} \sum_{i, l_i = p} \mathbf{x}_i$. $\mathbf{S}_I \in \mathbb{R}^{D \times D}$ and $\mathbf{S}_O \in \mathbb{R}^{D \times D}$ are the intra-class and out-of-class scatter matrices defined by:
\begin{equation}
\mathbf{S}_I = \sum_{i, l_i = p} (\mathbf{x}_i - \bar{\mathbf{x}}_p)(\mathbf{x}_i - \mathbf{x}_p)^T = \mathbf{X} \mathbf{L}_I \mathbf{X}^T,
\end{equation}
\begin{equation}
\mathbf{S}_O = \sum_{i, l_i \neq p} (\mathbf{x}_i - \bar{\mathbf{x}}_p)(\mathbf{x}_i - \mathbf{x}_p)^T = \mathbf{X} \mathbf{L}_O \mathbf{X}^T,
\end{equation}
where $\mathbf{L}_I = (1-\frac{2}{N_p} + \frac{1}{N_p^1}) \mathbf{e}_I \mathbf{e}_I^T$ and $\mathbf{L}_O = \mathbf{e}_O\mathbf{e}_O^T - \frac{1}{N_p}\mathbf{e}_I\mathbf{e}_O^T - \frac{1}{N_p}\mathbf{e}_I\mathbf{e}_O^T + \frac{1}{N_p}\mathbf{e}_I\mathbf{e}_I^T$.

The optimal projection matrix is obtained by solving the trace ratio problem \cite{jia2009trace} defined on $\mathbf{S}_I$ and $\mathbf{S}_O$ and is formed by the eigen-vectors of the matrix $\mathbf{S} = \mathbf{S}_I^{-1} \mathbf{S}_O$ corresponding to the $d_p$ maximal eigen-values. By assuming that the number of images depicting person $p$ is smaller than the number of the images depicting all other persons in the facial image database (which is usually the case), the rank of $\mathbf{S}$ is equal to $N_p - 1$. Thus, the maximal dimensionality of the class-specific discriminant space is equal to $d_p = min(N_p-1,D)$.

\subsection{Nonlinear case based on kernels}\label{SS:OriginalKernelCSLDA}
In order to define a non-linear mapping $f(\cdot)$, Class-Specific Kernel Discriminant Analysis \cite{Goudelis2007} applies a two-step process; the input space $\mathbb{R}^D$ is first non-linearly mapped to the so-called kernel space $\mathcal{F}$ using a function $\phi(\cdot)$, so that:
\begin{equation}
\mathbf{x}_i \in \mathbb{R}^D \:\overset{\phi(\cdot)}{\Rightarrow}\: \phi(\mathbf{x}_i) \in \mathcal{F}.
\end{equation}
Then, a linear mapping $\mathbf{W}_{\phi} \in \mathbb{R}^{|\mathcal{F}| \times d_p}$ is obtained by minimizing the intra-class and out-of-class distances defined as follows:
\begin{equation}
D_I = \sum_{i, l_i = p} \| \mathbf{W}_{\phi}^T \phi(\mathbf{x}_i) - \bar{\mathbf{x}}_p^{\phi} \|_2^2 = tr( \mathbf{W}_{\phi}^T \mathbf{S}_I^{\phi} \mathbf{W}_{\phi} )
\end{equation}
and
\begin{equation}
D_O = \sum_{i, l_i \neq p} \| \mathbf{W}_{\phi}^T \phi(\mathbf{x}_i) - \bar{\mathbf{x}}_p^{\phi} \|_2^2 = tr( \mathbf{W}_{\phi}^T \mathbf{S}_O^{\phi} \mathbf{W}_{\phi} ),
\end{equation}
where $\bar{\mathbf{x}}_p^{\phi} = \frac{1}{N_p} \sum_{i, l_i = p} \phi(\mathbf{x}_i)$. Here, the intra-class and out-of-class scatter matrices are defined in the kernel space $\mathcal{F}$ by:
\begin{equation}
\mathbf{S}_I^{\phi} = \sum_{i, l_i = p} (\phi(\mathbf{x}_i) - \bar{\mathbf{x}}_p^{\phi})(\phi(\mathbf{x}_i) - \bar{\mathbf{x}}_p^{\phi})^T = \mathbf{\Phi} \mathbf{L}_I \mathbf{\Phi}^T,
\end{equation}
\begin{equation}
\mathbf{S}_O^{\phi} = \sum_{i, l_i \neq p} (\phi(\mathbf{x}_i) - \bar{\mathbf{x}}_p^{\phi})(\phi(\mathbf{x}_i) - \bar{\mathbf{x}}_p^{\phi})^T = \mathbf{\Phi} \mathbf{L}_O \mathbf{\Phi}^T.
\end{equation}
$\mathbf{\Phi} = [\phi(\mathbf{x}_1),\dots,\phi(\mathbf{x}_N)] \in \mathbb{R}^{|\mathcal{F}| \times N}$ is a matrix having as columns the training data representations in $\mathcal{F}$. The Representer Theorem \cite{Scholkopf2001learning} states that the linear mapping in $\mathcal{F}$ can be expressed as a linear combination of the training data representation, i.e.:
\begin{equation}\label{Eq:RepresenterTheorem}
\mathbf{W}_{\phi} = \mathbf{\Phi} \mathbf{A},
\end{equation}
where $\mathbf{A} \in \mathbb{R}^{N \times d_p}$. Using (\ref{Eq:RepresenterTheorem}), we obtain $D_I = tr( \mathbf{A}^T \mathbf{K} \mathbf{L}_I \mathbf{K} \mathbf{A} )$ and $D_O = tr( \mathbf{A}^T \mathbf{K} \mathbf{L}_O \mathbf{K} \mathbf{A} )$, where $\mathbf{K} = \mathbf{\Phi}^T \mathbf{\Phi}$ is the so-called kernel matrix.

Two solutions have been proposed in order to obtain the optimal matrix $\mathbf{A}$. The first, applies eigenanalysis to the matrix $(\mathbf{K}\mathbf{L}_I\mathbf{K})^{-1}(\mathbf{K}\mathbf{L}_O\mathbf{K})$ and forms $\mathbf{A}$ with the eigen-vectors corresponding to the $d_p$ maximal eigen-values \cite{Goudelis2007,Iosifidis2015CSRDA}. The second one, noted as Class-Specific Kernel Spectral Regression, applies a two-step process; eigenanalysis of the matrix $\mathbf{L}_I^{-1}\mathbf{L}_O$ in order to obtain the eigen-vectors corresponding to the $d_p$ maximal eigen-values, i.e. $\mathbf{T} = [\mathbf{t}_1,\dots,\mathbf{t}_{d_p}]$, and solution of a kernel regression problem given by $\mathbf{A} = \mathbf{K}^{-1} \mathbf{T}$ \cite{Arashloo2014class}. In \cite{Arashloo2014class} it has been also shown that the eigenanalysis of $\mathbf{L}_I^{-1}\mathbf{L}_O$ can be readily obtained by applying a fast matrix decomposition process. Based on this, an approximate solution has also been proposed in \cite{Iosifidis2016scalling}, where the kernel regression step was replaced by reduced kernel-based regression.

\section{Class-Specific Regression}\label{S:Proposed}
In this Section, we show that class-specific subspace learning is equivalent to a regression problem. We start by showing that the linear version of Class-Specific Discriminant Analysis is equivalent to linear regression using class-specific targets. Subsequently, we show that the approximate kernel-based version of Class-Specific Discriminant Analysis can be obtained by applying reduced kernel-based regression, generalizing our previous results in \cite{Iosifidis2016scalling,Iosifidis2017CSKDAr} for the case where a reduced reference vector set is used for kernel-based learning. Please note that the above analysis shows that the class specific kernel regression in \cite{Iosifidis2017CSKDAr} is equivalent to the class-specific kernel spectral regression in \cite{Iosifidis2016scalling} in both the cases where standard and reduced kernels are used. Moreover, we propose a new solution to the CSDA problem based on neural networks at the end of this section.

\subsection{Linear case}\label{SS:ProposedLinear}
Let us assume that the training vectors are centered with respect to $\bar{\mathbf{x}}_p$\footnote{This can always be done by using $\mathbf{X} \leftarrow \mathbf{X} - \frac{1}{N_p} \mathbf{X}\mathbf{e}_I\mathbf{e}^T$.}. Then, the intra-class and out-of-class scatter matrices are given by $\mathbf{S}_I = \mathbf{X} \mathbf{E}_I \mathbf{X}^T$ and $\mathbf{S}_O = \mathbf{X} \mathbf{E}_O \mathbf{X}^T$, respectively. We also define the matrix $\mathbf{S}_T = \mathbf{X} \mathbf{X}^T$ denoting the total scatter of the training data with respect to $\bar{\mathbf{x}}_p$. It is easy to show that $\mathbf{S}_T = \mathbf{S}_I + \mathbf{S}_O$. The optimal projection matrix is obtained by maximizing:
\begin{equation}\label{Eq:CriterionLinearProp}
\tilde{\mathcal{J}}(\mathbf{W}) = \mathcal{J}(\mathbf{W}) + 1 = \frac{ tr( \mathbf{W}^T \mathbf{S}_O \mathbf{W} ) }{ tr( \mathbf{W}^T \mathbf{S}_I \mathbf{W} ) } + 1  = \frac{ tr( \mathbf{W}^T \mathbf{S}_T \mathbf{W} ) }{ tr( \mathbf{W}^T \mathbf{S}_I \mathbf{W} ) }.
\end{equation}
Thus, $\mathbf{W}$ is obtained by applying eigenanalysis to the matrix $\mathbf{S} = \mathbf{S_T}^{-1}\mathbf{S}_I$, i.e. by solving the following problem:
\begin{equation}\label{Eq:LinearEigCSLDA}
\mathbf{X} \mathbf{E}_I \mathbf{X}^T \mathbf{w} = \lambda \mathbf{X}\mathbf{X}^T \mathbf{w}, \:\:\lambda \neq 0.
\end{equation}

Let us now consider a linear regression problem using target vectors $\mathbf{T} = [\mathbf{t}_1,\dots,\mathbf{t}_{d_p}]$, i.e.:
\begin{equation}\label{Eq:CriterionLinearPropRegr}
\hat{\mathcal{J}}(\mathbf{W}) = \|\mathbf{W}^T \mathbf{X} - \mathbf{T}\|_F^2.
\end{equation}
Let us also express the data projection matrix as a product of two matrices $\mathbf{W} = \mathbf{Q} \mathbf{R}$, where $\mathbf{Q} \in \mathbb{R}^{D \times d_p}$ and $\mathbf{R} \in \mathbb{R}^{d_p \times d_p}$. Then, we have:
\begin{equation}\label{Eq:LinearRegressionProblem}
\hat{\mathcal{J}} = \|\mathbf{R}^T \mathbf{Q}^T \mathbf{X} - \mathbf{T}\|_F^2.
\end{equation}
The saddle point of $\hat{\mathcal{J}}$ with respect to $\mathbf{R}$ is given for $\mathbf{R} = (\mathbf{Q}^T \mathbf{X}\mathbf{X}^T \mathbf{Q})^{-1} \mathbf{Q}^T \mathbf{X} \mathbf{T}^T$. Substituting $\mathbf{R}$ in (\ref{Eq:LinearRegressionProblem}), we obtain:
\begin{eqnarray}
\hat{\mathcal{J}} &=& \|\mathbf{T} \mathbf{X}^T \mathbf{Q} (\mathbf{Q}^T \mathbf{X}\mathbf{X}^T \mathbf{Q})^{-1} \mathbf{Q}^T \mathbf{X} - \mathbf{T}\|_F^2 \nonumber \\
&=& c - 2tr( (\mathbf{Q}^T \mathbf{X}\mathbf{X}^T \mathbf{Q})^{-1} (\mathbf{Q}^T \mathbf{X} \mathbf{T}^T \mathbf{T}\mathbf{X}^T \mathbf{Q}).
\end{eqnarray}
Thus, the solution of $\hat{\mathcal{J}}$ is given by solving the following problem:
\begin{equation}\label{Eq:LinearEigRegression}
\mathbf{X} \mathbf{T}^T \mathbf{T} \mathbf{X}^T \mathbf{w} = \lambda \mathbf{X}\mathbf{X}^T \mathbf{w}, \:\:\lambda \neq 0.
\end{equation}

By comparing (\ref{Eq:LinearEigRegression}) with (\ref{Eq:LinearEigCSLDA}) we observe that the solution of the linear version of Class-Specific Discriminant Analysis is equivalent to a linear regression problem, where the target vectors satisfy $\mathbf{T}^T \mathbf{T} = \mathbf{E}_I$. We will show how to calculate such target vectors in Subsection \ref{SS:ProposedKernelCSLDA}.

\subsection{Nonlinear case based on kernels}\label{SS:ProposedKernelCSLDA}
Similar to the linear case, we assume that the training vectors are centered with respect to $\bar{\mathbf{x}}_p^{\phi}$\footnote{This can always be done by centering the kernel matrix $\mathbf{K}$ with respect to $\frac{1}{N_p}\mathbf{K}\mathbf{e}_I$. Test kernel vectors should be centered accordingly.}. Then, the intra-class and out-of-class scatter matrices expressed in $\mathcal{F}$ are given by $\mathbf{S}_I^{\phi} = \mathbf{\Phi} \mathbf{E}_I \mathbf{\Phi}^T$ and $\mathbf{S}_O^{\phi} = \mathbf{\Phi} \mathbf{E}_O \mathbf{\Phi}^T$, respectively. In addition, we define the matrix $\mathbf{S}_T^{\phi} = \mathbf{\Phi} \mathbf{\Phi}^T = \mathbf{S}_I^{\phi} + \mathbf{S}_O^{\phi}$ denoting the total scatter of the training data in $\mathcal{F}$ with respect to $\bar{\mathbf{x}}_p^{\phi}$.

Let us express the data projection matrix in $\mathcal{F}$ as a linear combination of $K$ reference vectors $\mathbf{\Psi} \in \mathbb{R}^{|\mathcal{F}| \times K}$, i.e.:
\begin{equation}\label{Eq:ReducedReprTh}
\mathbf{W}_{\phi} = \mathbf{\Psi} \mathbf{A},
\end{equation}
where $\mathbf{A} \in \mathbb{R}^{K \times d_p}$. The optimal $\mathbf{A}$ is obtained by maximizing:
\begin{eqnarray}
\tilde{\mathcal{J}}(\mathbf{A}) &=& \mathcal{J}(\mathbf{A}) + 1 = \frac{ tr( \mathbf{A}^T \mathbf{\Psi}^T \mathbf{S}_O^{\phi} \mathbf{\Psi} \mathbf{A} ) }{ tr( \mathbf{A}^T \mathbf{\Psi}^T \mathbf{S}_I^{\phi} \mathbf{\Psi} \mathbf{A} ) } + 1  \nonumber \\
&=& \frac{ tr( \mathbf{A}^T \mathbf{\Psi}^T \mathbf{S}_T^{\phi} \mathbf{\Psi} \mathbf{A} ) }{ tr( \mathbf{A}^T \mathbf{\Psi}^T \mathbf{S}_I^{\phi} \mathbf{\Psi} \mathbf{A} ) } \nonumber \\
&=& \frac{ tr( \mathbf{A}^T \mathbf{\Psi}^T \mathbf{\Phi} \mathbf{\Phi}^T \mathbf{\Psi} \mathbf{A} ) }{ tr( \mathbf{A}^T \mathbf{\Psi}^T \mathbf{\Phi} \mathbf{E}_I \mathbf{\Phi}^T \mathbf{\Psi} \mathbf{A} ) } \nonumber \\
&=& \frac{ tr( \mathbf{A}^T \tilde{\mathbf{K}} \tilde{\mathbf{K}}^T \mathbf{A} ) }{ tr( \mathbf{A}^T \tilde{\mathbf{K}} \mathbf{E}_I \tilde{\mathbf{K}}^T \mathbf{A} ) }.
\end{eqnarray}
Thus, $\mathbf{A}$ is obtained by applying eigenanalysis to the matrix $(\tilde{\mathbf{K}} \tilde{\mathbf{K}}^T)^{-1} (\tilde{\mathbf{K}} \mathbf{E}_I \tilde{\mathbf{K}}^T)$, i.e. by solving the following problem:
\begin{equation}\label{Eq:KernelEigCSLDA}
\tilde{\mathbf{K}} \mathbf{E}_I \tilde{\mathbf{K}}^T \mathbf{a} = \lambda \tilde{\mathbf{K}} \tilde{\mathbf{K}}^T \mathbf{a}, \:\:\lambda \neq 0.
\end{equation}

Next, we consider a linear regression problem in $\mathcal{F}$ using target vectors $\mathbf{T} = [\mathbf{t}_1,\dots,\mathbf{t}_{d_p}]$, i.e.:
\begin{equation}
\hat{\mathcal{J}} = \|\mathbf{W}_{\phi}^T \mathbf{\Phi} - \mathbf{T}\|_F^2 = \|\mathbf{A}^T \tilde{\mathbf{K}} - \mathbf{T}\|_F^2.
\end{equation}
where we have also exploited (\ref{Eq:ReducedReprTh}).

Similar to the linear case, we set $\mathbf{A} = \mathbf{Q} \mathbf{R}$, where $\mathbf{Q} \in \mathbb{R}^{K \times d_p}$ and $\mathbf{R} \in \mathbb{R}^{d_p \times d_p}$. Then, we have:
\begin{equation}\label{Eq:KernelRegressionProblem}
\hat{\mathcal{J}} = \|\mathbf{R}^T \mathbf{Q}^T \tilde{\mathbf{K}} - \mathbf{T}\|_F^2.
\end{equation}
The saddle point of $\hat{\mathcal{J}}$ with respect to $\mathbf{R}$ is given for $\mathbf{R} = (\mathbf{Q}^T \tilde{\mathbf{K}}\tilde{\mathbf{K}}^T \mathbf{Q})^{-1} \mathbf{Q}^T \tilde{\mathbf{K}} \mathbf{T}^T$. Substituting $\mathbf{R}$ in (\ref{Eq:KernelRegressionProblem}), we obtain:
\begin{eqnarray}
\hat{\mathcal{J}} &=& \|\mathbf{T} \tilde{\mathbf{K}}^T \mathbf{Q} (\mathbf{Q}^T \tilde{\mathbf{K}}\tilde{\mathbf{K}}^T \mathbf{Q})^{-1} \mathbf{Q}^T \tilde{\mathbf{K}} - \mathbf{T}\|_F^2 \nonumber \\
&=& c - 2tr( (\mathbf{Q}^T \tilde{\mathbf{K}}\tilde{\mathbf{K}}^T \mathbf{Q})^{-1} (\mathbf{Q}^T \tilde{\mathbf{K}} \mathbf{T}^T \mathbf{T}\tilde{\mathbf{K}}^T \mathbf{Q}).
\end{eqnarray}
Thus, the solution of $\hat{\mathcal{J}}$ is given by solving for:
\begin{equation}\label{Eq:KernelEigRegression}
\tilde{\mathbf{K}} \mathbf{T}^T \mathbf{T} \tilde{\mathbf{K}}^T \mathbf{q} = \lambda \tilde{\mathbf{K}}\tilde{\mathbf{K}}^T \mathbf{q}, \:\:\lambda \neq 0.
\end{equation}

By comparing (\ref{Eq:KernelEigRegression}) with (\ref{Eq:KernelEigCSLDA}) we observe that the solution of the approximate kernel Class-Specific Discriminant Analysis \cite{Iosifidis2016scalling} is equivalent to a reduced kernel regression problem, where the target vectors satisfy $\mathbf{T}^T \mathbf{T} = \mathbf{E}_I$. This is not surprising, since the kernel-based solution is obtained by applying the method described in subsection \ref{SS:ProposedLinear} in $\mathcal{F}$.

When the training vectors are used as reference vectors, i.e. when $\mathbf{\Psi} = \mathbf{\Phi}$, the above analysis shows that the Class-Specific Kernel Discriminant Analysis method is equivalent to (low-rank) kernel regression, which is the case of \cite{Iosifidis2017CSKDAr}. When a reduced kernel is used, the above analysis is equivalent to Approximate Class-Specific Kernel Discriminant Analysis (ACSKDA) \cite{Iosifidis2016scalling}. However, here we should note that while in ACSKDA the analysis involves the intra-class and out-of-class scatter matrices, in the above analysis the intra-class and total scatter matrices are used. Moreover, as has been shown in \cite{Iosifidis2016scalling}, where a Spectral Regression process is used, reference vectors can be defined by using a subset of the training vectors, or by applying clustering on the training data and using the cluster centers. This case corresponds to an approximate solution of the original Class-Specific Kernel Discriminant Analysis. We have observed that the use of cluster centers, e.g. obtained by applying $K$-Means to the training vectors, as reference vectors provides good performance, when compared to other alternatives \cite{Iosifidis2016scalling}.

Target vectors used in both linear and non-linear case can be calculated by applying an efficient orthogonalization technique exploiting the (class-specific) labels of the training data, as we have shown in our previous work \cite{Iosifidis2017CSKDAr}. This process is illustrated in Pseudocode \ref{EigenSolution}.

\section{Neural Class-Specific Regression}\label{SS:ProposedNetworkCSLDA}
\begin{figure*}[h!]
\centering \centerline{
\includegraphics[width=0.9\linewidth]{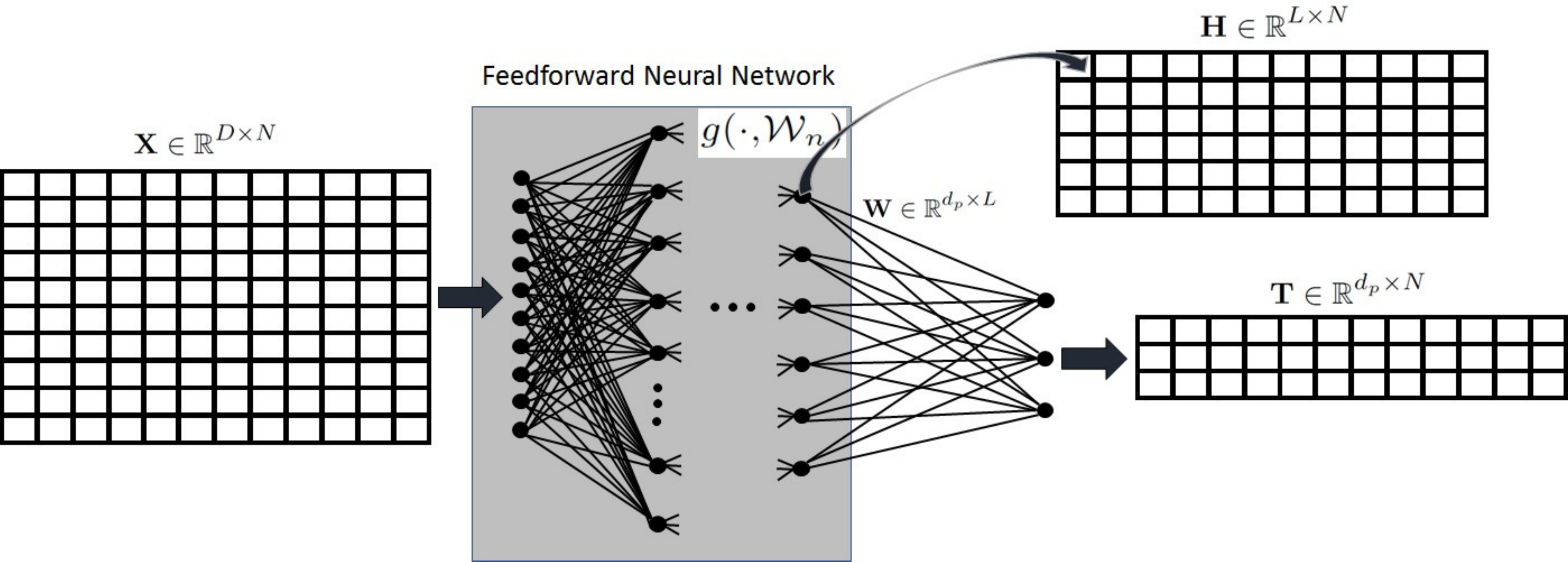} }
\caption{\it Neural network-based class-specific regression. The neural network is trained by using the facial image vectors $\mathbf{x}_i \in \mathbb{R}^D$ and the class-specific target vectors $\mathbf{t}_i \in \mathbb{R}^{d_p}$ forming the matrices $\mathbf{X}$ and $\mathbf{T}$, respectively. After training, the network maps an input vector $\mathbf{x}_i$ to $\mathbf{h}_i \in \mathbb{R}^L$ using the nonlinear function $g(\cdot,\mathcal{W}_n)$. A linear mapping $\mathbf{W} \in \mathbb{R}^{d_p \times L}$ is subsequently used to linearly map $\mathbf{h}_i$ to the corresponding class-specific representation in $\mathbb{R}^{d_p}$.}\label{fig:networkCSDA}
\end{figure*}

As has been shown above, both linear and approximate kernel Class-Specific Discriminant Analysis approaches are equivalent to linear regression problems in the $\mathbb{R}^D$ and $\mathcal{F}$, respectively, using the same target vectors defined based on the (class-specific) training labels (Pseudocode \ref{EigenSolution}). In order to derive a neural network based solution, let us define (with some abuse of notation) a non-linear mapping from the input space $\mathbb{R}^D$ to a feature space $\mathbb{R}^L$ obtained by applying
a non-linear function $g(\mathbf{x}_i, \mathcal{W}_n)$, such that:
\begin{equation}
\mathbf{x}_i \in \mathbb{R}^D \:\overset{g(\cdot,\mathcal{W}_n)}{\Rightarrow}\: \mathbf{h}_i \in \mathbb{R}^L.
\end{equation}
After mapping the training data in $\mathbb{R}^L$, a linear projection can be obtained by solving the CSDA problem (\ref{Eq:CriterionLinearProp}), or its equivalent class-specific regression problem (\ref{Eq:CriterionLinearPropRegr}), as illustrated in Figure \ref{fig:networkCSDA}. That is, the neural network-based class-specific mapping is obtained by minimizing:
\begin{equation}
\mathcal{J}(\mathbf{W},\mathcal{W}_n) = \|\mathbf{W}^T \mathbf{H} - \mathbf{T}\|_F^2,
\end{equation}
where $\mathbf{H}$ is a function of $\mathcal{W}_n$, i.e. $\mathbf{H} = g(\mathbf{X},\mathcal{W}_n)$.

The parameters of the above-described class-specific neural network are initialized randomly and can be optimized as follows:
\begin{itemize}
    \item \textit{Batch-based optimization:} In this case, the entire training set $\mathbf{X}$ is fed to the network in order to obtain the data representations in $\mathbb{R}^L$ and the optimal linear projection matrix $\mathbf{W}$ for the epoch $t$ is, subsequently calculated by:
        \begin{equation}
            \mathbf{W}_{(t)} = \mathbf{H}_{(t)}^\dag \mathbf{T}^T,
        \end{equation}
        where the symbol $^\dag$ denotes the pseudo-inverse of a matrix. The training error, then it is used in order to update the parameters of the network $\mathcal{W}_n$, based on gradient descent. Multiple training epoches are applied using the above-described process.
    \item \textit{Mini batch-based optimization:} In this case, we regard the entire process as a neural network having nonlinear activation functions in all layers, except the last one which is formed by linear neurons. Thus, the optimization of both the network's parameters $\mathcal{W}_n$ and the linear projection $\mathbf{W}$ can be performed sequentially, following mini batch-based gradient descent optimization.
\end{itemize}

We have found in our preliminary experiments that the latter approach leads to much faster solutions achieving good performance, when compared to the first one, and we use it in all our experiments. Moreover, as will be describe in Section \ref{S:Experiments}, we jointly train the network parameters $\mathcal{W}_n$ for multiple class-specific problems. This approach greatly speeds up the training process.
\begin{pseudocode}
\caption{Calculation of $\mathbf{T}$}\label{EigenSolution}
\begin{algorithmic}[1]
\Procedure{$\mathbf{T}$ = targets$\_$calculation}{$\mathbf{l}$, $p$, $d_p$}
\\
\State $N = length(\mathbf{l});$
\State $\mathbf{T} = rand(2,d_p+1);$  \:\:\:\:\:\: $\mathbf{Z} = zeros(N,d_p+1);$
\State $\mathbf{f}_1 = find(\mathbf{l}==p);$  \:\:\:\:\:\:\:\: $\mathbf{f}_2 = find(\mathbf{l}\neq p);$
\State $\mathbf{Z}(\mathbf{f}_1,:) = repmat(T(1,:),length(\mathbf{f}_1),1);$
\State $\mathbf{Z}(\mathbf{f}_2,:) = repmat(T(2,:),length(\mathbf{f}_2),1);$
\State $\mathbf{Z}(:,1) = ones(N,1)/ \sqrt{N};$
\State $\mathbf{M} = qr(\mathbf{Z});$ \:\:\:$\mathbf{M}(:,1) = [];$ \:\:\:$\mathbf{T} = \mathbf{M}^T$
\EndProcedure
\end{algorithmic}
\end{pseudocode}

\section{Discussion}\label{S:Discussion}
Here we provide discussion related to the properties of the class-specific regression models, compared to the original models based on eigenanalysis. We start by providing the time complexity of each variant. Subsequently, we discuss some limitations of the class-specific regression-based models and possible ways to address them.

In the linear case, the class-specific regression model involves the following processing steps:
\begin{itemize}
    \item Data centering, having a time complexity of $O(DN)$.
    \item Calculation of the target vectors $\mathbf{T}$, having a time complexity of $O(N_p N - \frac{1}{3} N_p^3)$ \cite{golub1996matrix}.
    \item Calculation of $\mathbf{W}$. In the case where a Cholesky decomposition-based solution is used, this step has a time complexity of $O(\frac{1}{6}D^3 + (N_p+N)D^2)$ \cite{Iosifidis2017CSKDAr}.
\end{itemize}
Thus, the overall time complexity of the linear class-specific regression model is $O(\frac{1}{6}D^3 + \frac{1}{6}D^3 + (N_p+N)D^2 + N_p N - \frac{1}{3} N_p^3 + DN)$.

The eigenanalysis based CSDA method involves the following processing steps:
\begin{itemize}
    \item Data centering, having a time complexity of $O(DN)$.
    \item Calculation of $\mathbf{S}_I$ and $\mathbf{S}_O$, having time complexity of $O(D^2N)$.
    \item Calculation of $\mathbf{S} = \mathbf{S}_I^{-1}\mathbf{S}_O$, having time complexity of $O(2D^3)$.
    \item Eigenanalysis of $\mathbf{S}$, having time complexity of $O(D^3)$.
\end{itemize}
Thus, the overall time complexity of CSDA is $O(3D^3 + D^2N + DN)$. Comparing the two approaches, we can see that both are linear with respect to the number of samples $N$ and cubic with respect to the data dimensionality $D$.

The time complexities of CS-KDA \cite{Arashloo2014class,Iosifidis2015CSRDA} is equal to $O(\frac{40}{3}N^3 + (D+d_p) N^2)$, while the time of the ACSKDA is equal to $O( (N_p^2+d_p+D)N + K^3 + d_p K^2 - \frac{1}{3}N_p^3)$ \cite{Iosifidis2016scalling}. As can be seen, by adopting an approximate kernel-based solution the time complexity becomes a cubic function of the number of reference vectors and positive samples $K$ and $N_p$, respectively. Regarding the time complexity of the neural network-based solution, it is a function of the number of parameters of the adopted architecture. However, by taking into account the high parallelization of feedforward networks, the time cost can be highly reduced.

One of the disadvantages of adopting a regression model is that, since such models optimize the mean square error with respect to the targets, the ratio between the cardinalities of the positive and negative classes is important. That is, in the case where the number of positive samples is much lower than the number of negative samples, the solution of the regression model will focus more on providing small training error on the negative class, while achieving a high error on the positive samples. In order to address this issue, weighted regression models can be adopted that increase the cost of training errors on the positive samples. Similar weighting schemes have also been used for eigenanalysis based discriminant analysis methods \cite{tang2005linear,li2009nonparametric}. A disadvantage of all class-specific models compared to their multi-class counterparts is related to their application in multi-class problems. In that case one needs to learn multiple models (in an one-versus-rest manner) increasing the overall computational cost linearly with respect to the number of classes. However, as will be discussed in subsection \ref{SS:ImplDetails}, in the case of class-specific regression models the overall computational cost can be highly reduced. Finally, one advantage of the proposed neural network-based class-specific model is the fact that it can easily extended in order to learn class-specific representations directly from (raw image) data, e.g. by including convolutional layers at the beginning of the architecture depicted in Figure \ref{fig:networkCSDA}.

\section{Experiments}\label{S:Experiments}
In this Section, we provide experimental results obtained by applying the regression-based CSDA methods described above on two face verification problems. First, we describe the two datasets used in our experiments. Later, we provide details on the experimental setup followed and speed up schemes we used in order to accelerate the training of the multiple class-specific models involved in each experiment.

\subsection{Datasets}\label{SS:Datasets}
\begin{figure}
[h!]
\centering{\includegraphics[width=0.9\linewidth]{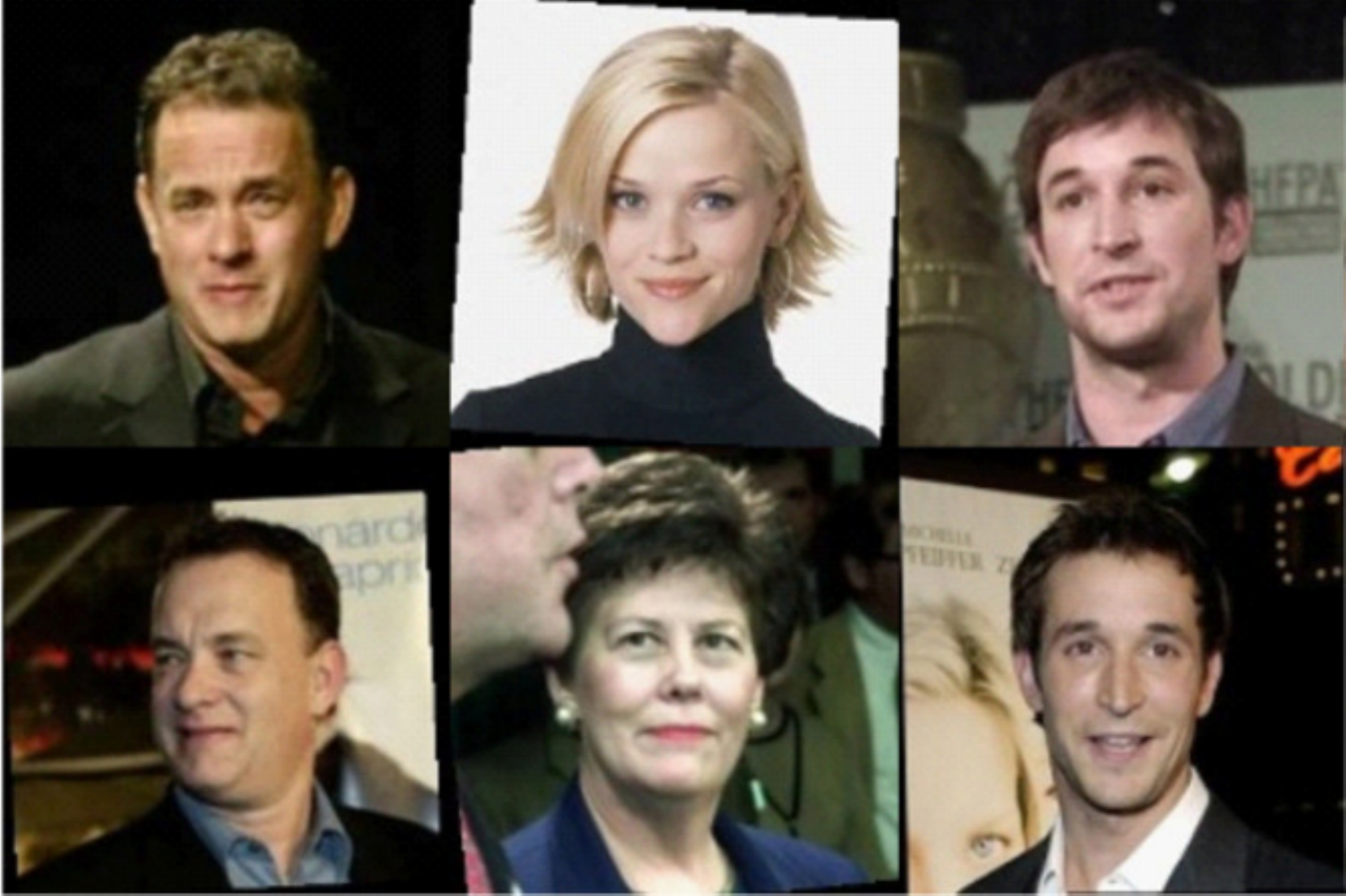}\\. \\ \includegraphics[width=0.9\linewidth]{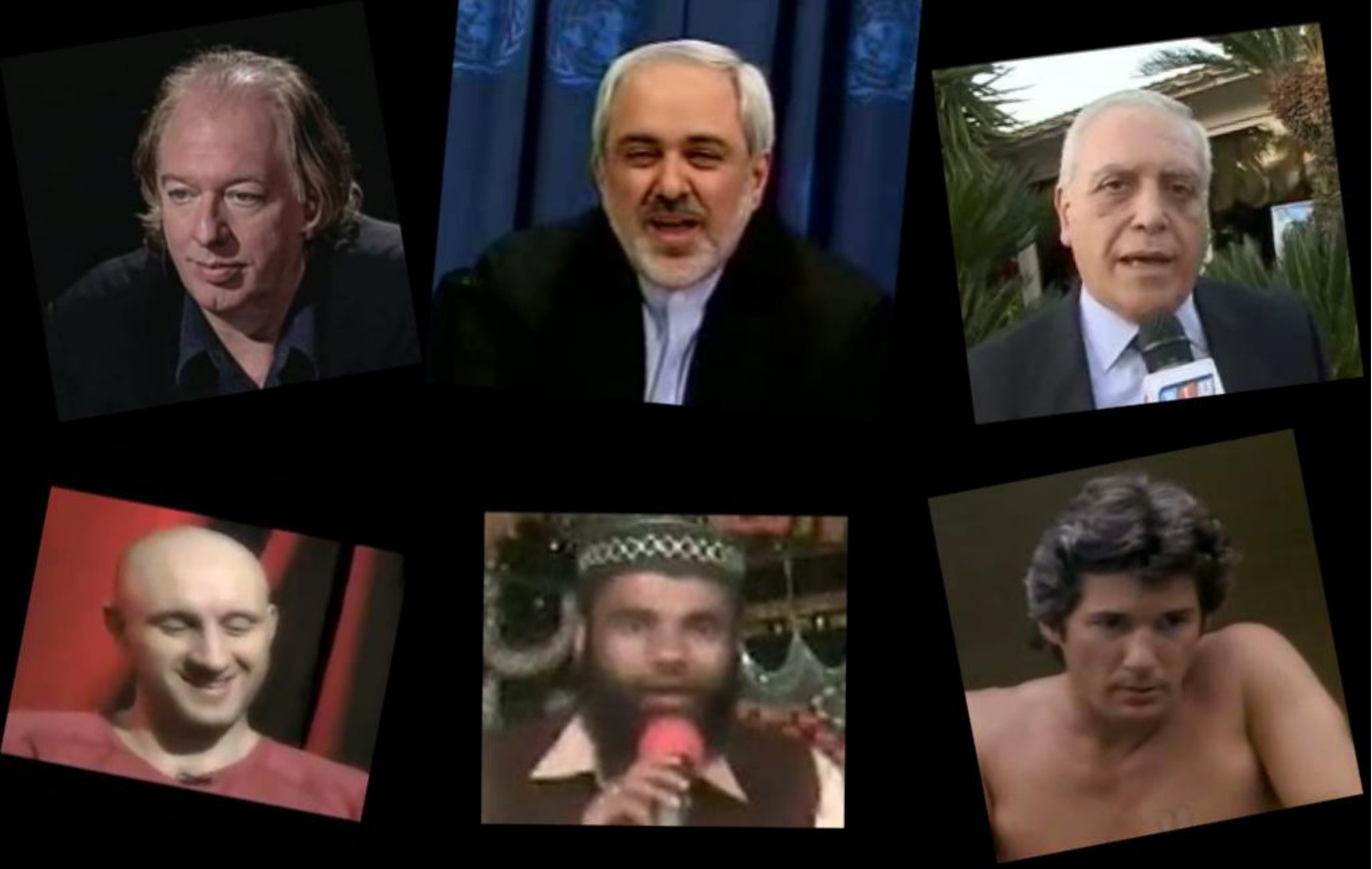}}                                                                                                                                  \caption{\it Facial images depicting persons from (top) PubFig$+$LFW and (bottom) YouTube Faces datasets.}\label{FacialImageDatasetsExamples}
\end{figure}

We have employed two facial image datasets, namely PubFig$+$LFW \cite{ortiza2014web} and Youtube Faces (YTFaces) \cite{wolf2011youtubeFaces}. The PubFig$+$LFW dataset describes a medium-scale facial image analysis problem. It is formed by the $47189$ facial images depicting $200$ persons coming from the Public Figures (PubFig) and the Labeled Faces in the Wild (LFW) datasets. The YTFaces dataset has been collected from YouTube. It is formed by $621126$ facial images depicting $1595$ person ID classes and it corresponds to a highly imbalanced problem. We kept the classes formed by at least $500$ images, leading to a dataset formed by $370319$ images depicting $340$ persons. Figure \ref{FacialImageDatasetsExamples} illustrates images from these two datasets.

\subsection{Experimental setup}\label{SS:ExpSetup}
On each of the datasets, we form multiple verification problems. That is, each ID class is split in two sets, one to be used for training and the remaining one for evaluation. On the the PubFig$+$LFW dataset we use the provided $75\%/25\%$ partition. Since there is no widely adopted dataset partitioning for single image-based verification on the YouTube dataset, we perform five experiments and on each experiment we use a random $70\%/30\%$ partition of each class. Here we should note that YouTube Faces recently has been used for face verification using image pairs, e.g. in \cite{wolf2011youtubeFaces}, however, in this paper we apply single-image verification. Facial images of the PubFig$+$LFW and YouTube Faces datasets are represented by using the facial image representations suggested in \cite{ortiza2014web} and \cite{wolf2011youtubeFaces}, respectively.

On each experiment, we solve $P$ verification problems ($P = 200$ for PubFig$+$LFW and $P = 340$ for YouTube Faces). For each verification problem $p$, we use the vectors representing the training facial images of class $p$ as positive samples and the vectors representing the training facial images of the rest of the classes in as negative samples. The class-specific discriminant (sub-)space is determined by applying each of the methods and the class representation in the discriminant (sub-)space $\bar{\mathbf{z}}_p$ is calculated. Subsequently, the representations of the test facial images of all classes in that discriminant (sub-)space $\mathbf{z}_j$ are calculated and their similarity to the class representation is calculated using $s_j = \|\mathbf{z}_j - \bar{\mathbf{z}}_p\|_2^{-1}$. Similarity values of all test images are sorted in a decreasing order and the equal error rate (EER) metric is calculated. The above-described process is repeated for all ID classes in the dataset and the performance of each method is measured by using the mean EER value and the corresponding standard deviation (over the multiple experiments).

\subsection{Benchmark methods}\label{SS:BenchmarkMethods}
We tested the performance of all three (linear, approximate kernel and neural network-based) class-specific regression models (Table \ref{tbl:Results2}). These models are referred to as LinCSDA, AK-CSDA and NN-CSDA, respectively. We also tested the performance obtained by applying the following methods: Support Vector Machine (SVM), Ridge Regression-based classification (RRC), LDA, CS-LDA, Extreme Learning Machine (ELM) \cite{huang2012relm}, Reduced Kernel Support Vector Machine (RKSVM) \cite{lee2007reduced}, Approximate Kernel Extreme Learning Machine (AKELM) \cite{iosifidis2015akelm,iosifidis2017NeucomAKELM} and Random Feature Regression (RFR) \cite{rahimi2007random}. For the non-linear methods using reference vectors, we applied multiple experiments using the reference vector set cardinalities of $K = \{500, 1000, 1500, 2000, 2500\}$ and report the best performance. For the class-specific approaches, we applied multiple experiments using discriminant (sub-)space dimensionality $d_p = d, \:p=1,\dots,P$ for $d = \{1,5,10\}$.

\subsection{Implementation details}\label{SS:ImplDetails}
In the neural network-based approach, we adopted two single hidden layer networks for both datasets. Similarly, we set the learning rate as $1e-7$, the minibatch size $n = 200 $, the number of epoches as $40$ and sigmoid function as the activation function throughout the networks on both datasets.
We experimented numerous topologies with the number of neurons $L=\{100,200,500,1000,1500\}$ using a set of NVIDIA Tesla K80 GPUs in a parallel setting, and report the best collective results based on the validation set in Table \ref{tbl:Results1}. On the PubFig+LFW dataset, the number of hidden neurons is $1500$ while that of hidden neurons on YouTube Faces is $1000$. We omitted the result with the number of dimensionality equal to $10$ in the subspace due to the memory constraints of the graphic cards.

On the YouTube Faces dataset we perform five experiments and report the average EER and the corresponding standard deviation over all experiments. We have observed that the five datasets provided by PubFig$+$LFW database correspond to the same $P = 2000$ face verification problems. For the linear methods, the standard deviation values in \cite{Iosifidis2016scalling} correspond to the deviation of the performance due to different training subsets ($N=10000$) employed. A comparison of these results and those obtained using all training data is provided in Table \ref{tbl:Results1}. For the (approximate) kernel-based methods, the provided standard deviations correspond to the deviation of the performance due to different reference vectors selected over the five experiments (for different random $K$-Means initializations).

In order to speed up the training process over the multiple verification problems of each experiment, we exploit the fact that the training samples of all verification problems are the same (what changes is the labels used in order to define the target vectors, as detailed in Section \ref{S:Proposed} and Pseudocode \ref{EigenSolution}). Let us denote by $\mathbf{T}_p$ the matrix formed by the target vectors used for the determination of the discriminant (sub-)space of class $p$. Then, the solution of the (class-specific) regression problem (\ref{Eq:CriterionLinearPropRegr}) is given by $\mathbf{W} = (\mathbf{X}\mathbf{X}^T)^{-1} \mathbf{X} \mathbf{T}_p^T$. That is, the matrix $\mathbf{X}^\dag = (\mathbf{X}\mathbf{X}^T)^{-1} \mathbf{X}$ is used in all verification problems and can be calculated once. Subsequently, the determination of all class-specific discriminant sub-spaces can be obtained by applying a matrix multiplication between the matrices $\mathbf{X}^\dag$ and $\mathbf{T}_p^T$. In a similar way, we cluster the training vectors once in order to define the reference vectors, and calculate the matrix $\tilde{\mathbf{K}}^\dag = (\tilde{\mathbf{K}}\tilde{\mathbf{K}}^T)^{-1} \tilde{\mathbf{K}}$, which is used for all approximate kernel class-specific discriminant spaces, only once. For the neural network-based CSDA, we use the same network for all verification models, based on the intuition that such a choice will lead to a better representation of the facial images in the (shared) feature space $\mathbb{R}^L$. We implemented this stacking the target vectors of all class-specific regression models during the training process. During evaluation, we use the class-specific representations as described earlier in this subsection in order to measure the performance of each method.

\subsection{Results}
The regression-based class-specific methods provide good performance in both face verification problems. Linear class-specific regression achieves better performance when compared to the multi-class linear discriminant and regression methods, and similar performance to SVM. Approximate class-specific regression \cite{Iosifidis2016scalling} (corresponding to the reduced kernel-based class-specific regression in Subsection \ref{SS:ProposedKernelCSLDA}) outperforms the related multi-class regression models. Class-specific regression based on neural networks achieves competitive performance in PubFig$+$LFW dataset, which corresponds to a medium-scale verification problem, while it outperforms all non-linear models in the large-scale verification problem of YouTube Faces dataset. We believe that this is due to cardinality of the data (we have observed that convergence of the networks was difficult for PubFig$+$LFW dataset). We observe that linear models achieve better performance in both datasets. This might be due to the adopted data representations (this might also be the reason why both \cite{wolf2011youtubeFaces} and \cite{ortiza2014web} use linear models in their experiments). For nonlinear methods, we notice that the neural networks provide a performance gain over the kernel methods when the dataset size is large.
\begin{table}[]
\caption{Performance (mean ERR \%) of linear and non-linear methods on PubFig$+$LFW and YouTube Faces datasets\label{tbl:Results2}}
\resizebox{.9\linewidth}{!}{
{\begin{tabular}{l|c|c}\hline
Method                                      &        PubFig$+$LFW         &       YouTube Faces         \\\hline
SVM                                         & \bf    7.43                 & \bf  1.68$\pm$0.09          \\
RRC                                         &       17.25                 &      21.9$\pm$1.34          \\
LDA                                         &       17.24                 &     28.27$\pm$0.66          \\
CS-LDA (d=1)                                &       15.63                 &     23.62$\pm$0.77          \\
CS-LDA (d=5)                                &        6.05                 & \bf  1.81$\pm$1.41          \\
CS-LDA (d=10)                               & \bf    5.89                 &      2.71$\pm$1.89          \\\hline
LinCSDA (d=1)                               &       17.24                 &     22.13$\pm$1.3           \\
LinCSDA (d=5)                               & \bf    9.34                 &      0.87$\pm$0.05          \\
LinCSDA (d=10)                              &        9.45                 & \bf  0.53$\pm$0.02          \\\hline\hline
ELM  \cite{Iosifidis2016scalling}           &     25.81$\pm$0.91 (K=2500) &     17.08$\pm$0.57 (K=1500) \\
RKSVM \cite{Iosifidis2016scalling}          &     18.95$\pm$0.46 (K=2500) &     15.44$\pm$0.55 (K=1500) \\
AKELM \cite{Iosifidis2016scalling}          &     17.21$\pm$0.98 (K=2000) &     12.02$\pm$0.57 (K=500)  \\
RFR \cite{Iosifidis2016scalling}            &     20.62$\pm$0.84 (K=2500) &     27.67$\pm$0.48 (K=2000) \\\hline
AK-CSDA (d=1) \cite{Iosifidis2016scalling}  &      18.4$\pm$0.72 (K=2500) &     18.92$\pm$0.57 (K=1000) \\
AK-CSDA (d=5) \cite{Iosifidis2016scalling}  & \bf 11.49$\pm$0.82 (K=2500) &     12.98$\pm$0.93 (K=500)  \\
AK-CSDA (d=10) \cite{Iosifidis2016scalling} &     11.65$\pm$1.09 (K=2500) & \bf  2.26$\pm$0.11 (K=2500) \\\hline\hline
NN-CSDA (d=1)                               &     17.94 (L=1500)           &     22.47$\pm$1.05 (L=1000)  \\
NN-CSDA (d=5)                               & \bf 12.31 (L=1500)           & \bf  1.96$\pm$0.39 (L=1000)  \\
NN-CSDA (d=10)                              &     12.38 (L=1500)           &            -                \\\hline
\end{tabular}}{}}
\end{table}
\begin{table}[]
\caption{Performance (mean ERR \%) of linear methods on PubFig$+$LFW dataset\label{tbl:Results1}}
{\begin{tabular}{l|c|c}\hline
Method       &    Using $N=10^4$ \cite{Iosifidis2016scalling}  & Using $N=35469$ \\\hline
SVM          &      7.86$\pm$0.43                              &       7.43      \\
RRC          &     17.48$\pm$0.15                              &      17.25      \\
LDA          &     15.61$\pm$0.27                              &      17.24      \\
CS-LDA       &     14.69$\pm$0.34                              &      15.63      \\\hline
\end{tabular}}{}
\end{table}

\section{Conclusions}\label{S:Conclusions}
In this paper, we showed that class-specific subspace learning is equivalent to a regression problem using class-specific target vectors. Based on that, we derived linear, reduced kernel and neural network-based class-specific regression models suited for large-scale learning problems. Interesting future research directions based on the derived solutions include the application of class-specific models directly on (raw) image/video data for representation learning, e.g. by using convolutional and recurrent neural layers and the investigation of class-specific representations obtained by using such learning schemes.

\end{document}